%% file: main.tex
\documentclass[10pt,twocolumn,letterpaper]{article}

\usepackage[pagenumbers]{cvpr} 
\usepackage{multirow}

\input{preamble}
\definecolor{cvprblue}{rgb}{0.21,0.49,0.74}
\usepackage[pagebackref,breaklinks,colorlinks,allcolors=cvprblue]{hyperref}
\usepackage{array}
\usepackage{soul}
\usepackage[table]{xcolor}
\definecolor{tab1}{rgb}{1.0, 0.85, 0.85}
\definecolor{tab2}{rgb}{1.0, 0.92, 0.80}
\definecolor{tab3}{rgb}{1.0, 1.0, 0.80}
\definecolor{tab4}{rgb}{0.85, 1.0, 0.85}
\definecolor{tab5}{rgb}{0.85, 0.92, 1.0}

\def\paperID{1428} 
\def\confName{CVPR}
\def\confYear{2026}

\title{
Robotic VLA Benefits from Joint Learning with Motion Image Diffusion
}

\author{
Yu Fang$^{1,2}$\thanks{Work done during an internship at Salesforce.} \quad
Kanchana Ranasinghe$^{1}$ \quad
Le Xue$^{1}$ \quad
Honglu Zhou$^{1}$ \quad
Juntao Tan$^{1}$ \quad \\
Ran Xu$^{1}$ \quad
Shelby Heinecke$^{1}$ \quad
Caiming Xiong$^{1}$ \quad
Silvio Savarese$^{1}$ \quad \\
Daniel Szafir$^{2}$ \quad
Mingyu Ding$^{2}$ \quad
Michael S. Ryoo$^{1}$ \quad
Juan Carlos Niebles$^{1}$
\\
$^{1}$Salesforce AI Research \\
$^{2}$University of North Carolina at Chapel Hill \\
{\href{https://vla-motion.github.io/}{https://vla-motion.github.io/}}
}

\begin{document}
\maketitle

\begin{abstract}
Vision-Language-Action (VLA) models have achieved remarkable progress in robotic manipulation by mapping multimodal observations and instructions directly to actions.
However, they typically mimic expert trajectories without predictive motion reasoning, which limits their ability to reason about what actions to take.
To address this limitation, we propose joint learning with motion image diffusion, a novel strategy that enhances VLA models with motion reasoning capabilities.
Our method extends the VLA architecture with a dual-head design: while the action head predicts action chunks as in vanilla VLAs, an additional motion head, implemented as a Diffusion Transformer (DiT), predicts optical-flow-based motion images that capture future dynamics.
The two heads are trained jointly, enabling the shared VLM backbone to learn representations that couple robot control with motion knowledge.
This joint learning builds temporally coherent and physically grounded representations without modifying the inference pathway of standard VLAs, thereby maintaining test-time latency.
Experiments in both simulation and real-world environments demonstrate that joint learning with motion image diffusion improves the success rate of $\pi$-series VLAs to 97.5\% on the LIBERO benchmark and 58.0\% on the RoboTwin benchmark, yielding a 23\% improvement in real-world performance and validating its effectiveness in enhancing the motion reasoning capability of large-scale VLAs.
\end{abstract}

\section{Introduction}
Vision-Language-Action (VLA) models have emerged as a powerful paradigm for generalist robotic manipulation, capable of performing diverse tasks across environments and embodiments~\cite{brohan2022rt, zitkovich2023rt,team2024octo, liu2024rdt, kim2024openvla}.
These models are typically built upon large-scale Vision-Language Models (VLM)~\cite{liu2023visual,bai2023qwen,touvron2023llama,karamcheti2024prismatic,beyer2024paligemma}, which provide strong perceptual and linguistic understanding and are further equipped with action experts that translate multimodal representations into robot actions.
Despite their impressive performance, current VLAs exhibit a fundamental limitation: they lack an explicit mechanism to reason about future dynamics, and largely rely on imitating expert trajectories.
This absence of motion-level reasoning restricts their temporal understanding and ultimately limits generalization to new tasks, scenes, and embodiments.

\begin{figure}[t!]
  \centering
   \includegraphics[width=0.85\linewidth]{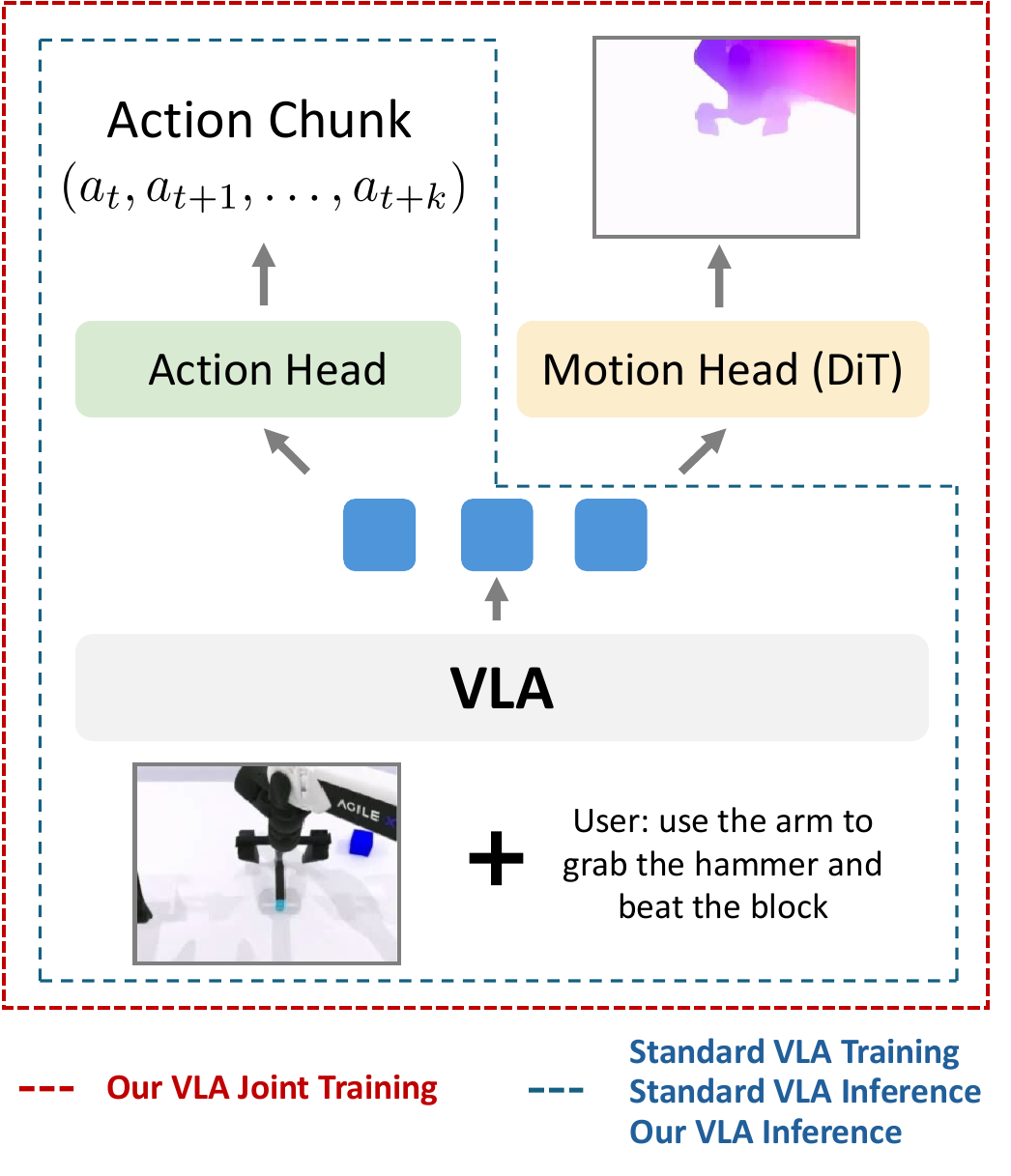}
   \caption{\textbf{Overview.} Our joint learning strategy seamlessly extends existing large-scale VLAs with motion image diffusion -- learning motion and action jointly through a shared VLM backbone, enhancing their motion reasoning abilities, and maintaining the same inference pipeline as in standard VLA models.}
   \label{fig:teaser}
\end{figure}

Motion understanding is a fundamental direction of effective visuomotor control.
In robotic manipulation, explicit motion representations, such as keypoint traces and optical flow, serve as popular abstractions for describing action-relevant information from manipulation videos~\cite{wen2023any, zheng2024tracevla, yang2025magma, li2025hamster, ranasinghe2025pixel}.
Recent studies have incorporated such representations into large-scale VLA models~\cite{belkhale2024rt, zheng2024tracevla, lee2025molmoact, yang2025magma}, demonstrating their potential to enhance motion awareness.
Meanwhile, world models and video prediction approaches implicitly describe motion through future image prediction based on current observations and have been introduced to guide policy learning~\cite{mendonca2023structured, black2023zero, du2023learning, hu2024video, bharadhwaj2024gen2act}.
Building upon this idea, recent works have explored unified VLA models~\cite{li2025unified,lyu2025dywa,zhu2025unified,wang2025unified,cen2025worldvla} that integrate action models with world models to provide predictive insights into how environments change over time.
However, these unified or video-generation-based architectures are often built independently from existing pretrained VLAs, making them difficult to integrate with existing high-performing VLA frameworks.
Moreover, while effective for capturing visual consistency, these methods largely overlook motion learning -- they learn to predict what the scene will look like, rather than how the robot should move.

To address these issues, we propose joint learning with motion image diffusion, a simple yet effective strategy that seamlessly improves VLAs with motion reasoning ability.
Specifically, as illustrated in Fig.~\ref{fig:teaser}, we extend VLA architecture with a dual-head design: an action head that predicts action chunks as in vanilla VLAs, and a motion head implemented as a Diffusion Transformer (DiT)~\cite{peebles2023scalable} that predicts optical-flow-based future motion images through diffusion.
Both heads share the same VLM backbone and are optimized jointly, enabling the model to learn temporally coherent and physically grounded representations that support both fine-grained control and motion understanding.
We find that optical-flow-based motion images offer an efficient and control-aligned supervision signal in joint learning.
Unlike future image prediction or language-based motion descriptions, optical flow directly encodes how the scene moves, making it inherently consistent with action learning.
This complementary supervision encourages the model to align physical motion dynamics with robot control, providing dense temporal guidance that improves visuomotor policy learning.
Importantly, our strategy integrates seamlessly into existing large-scale VLA models with no additional inference latency, making it practical for real-world robotic deployment.

Our key contributions are summarized as follows:

\begin{itemize}[leftmargin=*]
    \item We propose joint learning with motion image diffusion that seamlessly augments VLA models with motion reasoning capabilities, while preserving their real-time inference efficiency.

    \item We present motion image diffusion, implemented using a Diffusion Transformer (DiT), which provides dense pixel-level dynamic supervision that is complementary to sparse action supervision.
    We show that the optical-flow-based motion images are the most effective representation for joint action–motion learning.
    
    \item Extensive experiments demonstrate that our approach improves the performance of large-scale VLA models. For example, we enhance $\pi$-series VLA models to achieve an average success rate of 97.5\% on the LIBERO benchmark and 58.0\% on the RoboTwin benchmark.
\end{itemize}

\begin{figure*}[t]
  \centering
   \includegraphics[width=0.9\linewidth]{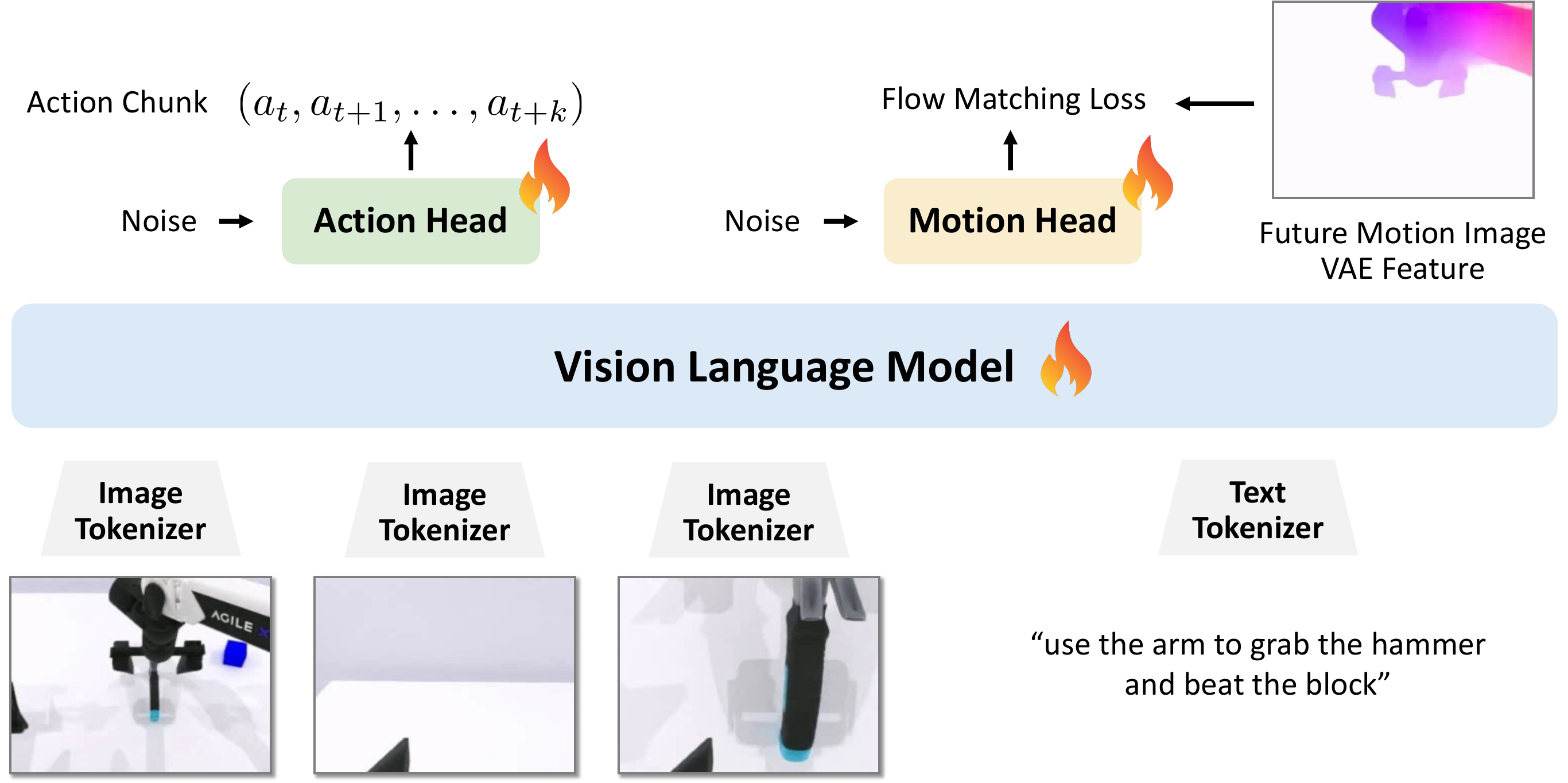}
   \caption{\textbf{Overview of joint learning VLA with motion image diffusion.}}
   \label{fig:method}
\end{figure*}

\section{Related Work}
\label{sec:related}
\textbf{Vision-Language-Action Models.}
With recent advances of multimodal Large Language Models (MLLM)~\cite{liu2023visual,bai2023qwen,touvron2023llama,karamcheti2024prismatic,beyer2024paligemma} and large-scale robotic datasets~\cite{ebert2021bridge, o2024open, khazatsky2024droid}, Vision-Language-Action (VLA) models have become a promising paradigm for generalist robotic manipulation~\cite{brohan2022rt, zitkovich2023rt, zhao2023learning,li2023vision, wu2023unleashing, belkhale2024rt, cheang2024gr, team2024octo, liu2024rdt, kim2024openvla,li2024cogact, wen2025tinyvla, qu2025spatialvla, liu2025towards, shi2025hi, liu2025hybridvla, black2024pi_0, intelligence2025pi_, bu2025agibot, reuss2025flower, song2025hume, pertsch2025fast}.
They are typically initialized from pretrained large models to inherit understanding and generalization abilities.
RT series~\cite{brohan2022rt, zitkovich2023rt, belkhale2024rt} pioneer the attempt to finetune MLLM for robotic manipulation, showing strong performance and generalization capability.
OpenVLA~\cite{kim2024openvla} scales VLA with a 7B model trained on 970k real-world demonstrations spanning diverse datasets.
$\pi$ series~\cite{black2024pi_0, intelligence2025pi_} adapt the PaliGemma~\cite{beyer2024paligemma} architecture with an action expert via flow matching~\cite{lipman2022flow}.
Despite the promising trend, these methods primarily focus on learning a direct mapping from language and observations to actions,
which lacks explicit motion learning.

\noindent \textbf{Motion Learning in VLA Models.}
Motion learning is a fundamental direction for understanding temporal dynamics in robotic manipulation~\cite{gao2023k, bharadhwaj2024track2act,cui2024dynamo, hu2024video, ranasinghe2025pixel, collins2025amplify, lin2024flowretrieval, fang2025kalm}.
Recently, several studies have explored incorporating motion as an additional modality into large-scale VLA frameworks~\cite{zhong2025flowvla, zheng2024tracevla}.
For example, FlowVLA~\cite{zhong2025flowvla} introduces optical flow images through visual Chain-of-Thought (CoT) reasoning for world model pre-training.
While this enhances motion reasoning, the flow signal is used only during pretraining and is not integrated into the robot policy, limiting its influence on downstream action generation.
Similarly, TraceVLA~\cite{zheng2024tracevla} and WorldVLA~\cite{cen2025worldvla} employ predictive modules that model object trajectories or future frames as auxiliary signals, but these methods focus on appearance-level reconstruction rather than motion-level dynamics.
Other frameworks based on video prediction~\cite{black2023zero, ko2023learning, Gu2023SeerLI, du2023learning, nair2022r3m, Guo2024PredictionWA, bharadhwaj2024gen2act, hu2024video, ye2024latent, wen2024vidman, Jeong2025ObjectCentricWM, feng2025vidar, Cheang2025GR3TR} learn latent dynamics for planning but do not couple them directly with VLA policies.
In contrast, our work emphasizes explicit motion learning through optical flow prediction, which enhances the learning of action generation.

\noindent \textbf{Unified Understanding and Generation Models.}
Recently, many MLLMs have introduced unifying visual understanding and visual generation, enabling large models to generate both language and visual content seamlessly~\cite{team2024chameleon,wang2024emu3,chen2025blip3,blip3onext, unigen}.
In the robotics domain, several papers have proposed unified architectures that generate images and actions~\cite{li2025unified,lyu2025dywa,zhu2025unified,wang2025learning,wang2025unified,cen2025worldvla}.
For example, UVA~\cite{li2025unified} proposes a unified architecture that generates images and actions with separate diffusion heads.
WorldVLA~\cite{cen2025worldvla} explores a discrete autoregressive unified model for both perception and action generation.
However, these unified architectures cannot be easily integrated into or benefit from high-performing VLA frameworks, and often fail to explore explicit motion representations.

\section{Method}
\label{sec:method}
\subsection{Preliminaries}
Modern Vision-Language-Action (VLA) models~\cite{black2024pi_0, intelligence2025pi_} typically consist of two major components: a pretrained Vision-Language Model (VLM) backbone for multimodal perception, and an action head for visuomotor control.
The VLM encodes visual observations $o_t$ and a language instruction $l$ into a shared multimodal representation:
\begin{equation}
z_t = \mathrm{VLM}(o_t, l),
\end{equation}
where $z_t$ captures multimodal information at time $t$.
On top of this backbone, the action head (also referred to as an action expert) adapts the VLM to robotic control, by predicting a chunk of $k$ future actions conditioned on $z_t$:
\begin{equation}
A_t = \pi_\theta(z_t),
\end{equation}
where $A_t = a_{t:t+k} \in \mathbb{R}^{k \times d}$ and $d$ denote the robot's degrees of freedom.
VLA architecture allows the policy to directly map multimodal knowledge to actions, but it lacks explicit modeling of motion and temporal dynamics. 

\subsection{Joint Learning with Motion Image Diffusion}
\label{sec:method_jointlearning}
In this paper, we argue that action learning and motion learning are inherently complementary: actions describe how the robot moves in physical space, while motion represents how the scene changes in pixel space.
To enhance the motion reasoning capability of a pretrained VLA model, as illustrated in Fig.~\ref{fig:method}, we propose a joint learning strategy with motion image diffusion.
Specifically, based on a standard pretrained VLA architecture, we propose to integrate an additional motion head $\mu_\psi$ that operates in parallel with the action head $\pi_\theta$.
Both heads share the same multimodal representation $z_t$ produced by the VLM backbone.
The motion head predicts latent motion tokens $m_t$ in a diffusion process conditioned on $z_t$:
\begin{equation}
    m_t = \mu_\psi (z_t).
\end{equation}
The latent motion tokens are subsequently decoded into a motion image $M_t = f(m_t)$ via a frozen VAE decoder $f$, which provides a compact and stable latent space.
Inspired by unified MLLM architectures for image understanding and generation~\cite{chen2025blip3}, we implement $\mu_\psi$ as a Diffusion Transformer (DiT)~\cite{peebles2023scalable} that models the conditional denoising process in the latent space.
Hence, this parallel design adapts the VLM backbone to learn two complementary forms of transformation: while $\pi_\theta$ learns to generate future actions in control space, $\mu_\psi$ learns to model the corresponding pixel-space dynamics, enabling the VLM backbone to encode temporally coherent visuomotor representations.

During training, following prior diffusion-based visuomotor policies~\cite{chi2025diffusion, black2024pi_0, intelligence2025pi_}, we adopt the flow matching loss~\cite{lipman2022flow} for both heads.
In this way, each head learns a time-dependent vector field that transforms a noisy sample into the ground truth target.
Specifically, for a general target signal $X_t \in \{A_t, m_t\}$, we define a noisy sample by linearly interpolating between the clean target $X_t$ and a Gaussian noise $\epsilon \sim \mathcal{N}(0, I)$ of the same shape:
\begin{equation}
X_t^\tau = \tau X_t + (1-\tau) \epsilon
\end{equation}
where $\tau \in [0, 1]$ denotes a sampled interpolation weight from a Beta distribution.
The objective of each head is to predict a velocity field $v(X_t^\tau, o_t)$ that matches the target flow $u(X_t^\tau|X_t)$:
\begin{equation}
    \mathcal{L}_{\text{FM}}(X_t) = 
    \mathbb{E}_{p(X_t|o_t)}
    \Big[
    \big\|v(X_t^\tau, o_t) - u(X_t^\tau | X_t)\big\|_2^2
    \Big].
\end{equation}

We apply the same loss independently to both heads:
\begin{equation}
\begin{cases}
    \mathcal{L}_{\text{action}} = \mathcal{L}_{\text{FM}}(A_t) \\
    \mathcal{L}_{\text{motion}} = \mathcal{L}_{\text{FM}}(m_t),
\end{cases}
\end{equation}
and jointly optimize the entire model with the total objective:
\begin{equation}
    \mathcal{L} = \mathcal{L}_\text{action} + \mathcal{L}_\text{motion}
\end{equation}
By coupling these two complementary objectives through a shared VLM backbone, the model now develops a more structured temporal representation that enhances motion reasoning and leads to more consistent visuomotor control.

\subsection{Training Details}
\label{sec:training}
\noindent \textbf{Supervision with Optical Flow.}
For motion supervision, we adopt image-based optical flow (termed as flow image) as the motion representation.
We leverage RAFT~\cite{teed2020raft} to calculate ground-truth pixel motion between each pair of observations $(o_t, o_{t+k})$ in the training datasets.
Each pixel motion contains two channels for spatial directions, which is mapped to HSV space and converted into the 3-channel RGB space.
This leverages the benefits of the pretrained VLM workflow.
We ensure that the motion signal spans the same temporal window as the action chunk to align temporal consistency between the action and motion signals.
The calculated flow is converted into an RGB image of the same spatial resolution as the input observations, allowing pixel-level motion alignment with the visual inputs.

\noindent \textbf{Diffusing Motion in Latent Space.}
Directly diffusing full-resolution optical flow images is computationally expensive and often unstable due to the high dimensionality and pixel-level noise in motion representations.
To address this issue, we adopt a frozen VAE to encode the flow images into a compact latent space $M_t \in \mathbb{R}^{4\times H/8 \times W/8}$ before applying diffusion.
The motion head is trained to generate such motion tokens instead of raw flow images, which offers several advantages:
1) Diffusing in compact latent space reduces spatial redundancy and stabilizes diffusion-based denoising, preserving the real-time efficiency of VLA models.
2) Latent representation abstracts local pixel variations into structured motion features, allowing the model to focus on learning consistent temporal dynamics rather than pixel-level appearance noise.
3) Operating in the token space enables the motion head to share a consistent embedding scale and conditioning structure with the action head, facilitating joint optimization under the shared VLM backbone.
Notably, VAE is always frozen for all procedures.
During inference, the predicted motion token can be decoded into a visualized flow image without any optimization.

\noindent \textbf{Training Procedure.}
We adopt a two-stage training pipeline to ensure stable optimization of both the action and motion branches.
For pretraining the motion head, we leverage the DROID dataset~\cite{teed2020raft}, a large-scale real robot dataset, to produce optical flow from diverse real robotic environments.
This enables the model to acquire general motion priors independent of task-specific control behaviors.
We first attach the motion head in parallel with the pretrained action expert and freeze all other parameters.
Only the motion head is optimized during this warm-up phase, allowing for stable adaptation from learning static visual appearance to dynamic motion prediction.
After the warm-up phase, we unfreeze the full architecture, except for the VAE encoder and decoder, and conduct joint training of both heads, optimizing the combined flow-matching objective as described in Sec.~\ref{sec:method_jointlearning}.
This stage enables the shared VLM backbone to jointly align action and motion representations, facilitating motion-aware control learning.
The proposed procedure is compatible with a wide range of VLA architectures, requiring only minimal modifications to integrate the additional motion head.

\section{Experiments}
\input{tabs/libero}
\subsection{Implementation Details}
In our experiments, we initialize both the VLM backbone and the action head from pretrained VLA checkpoints to retain strong visuomotor priors.
For example, developed upon the $\pi$-series VLAs~\cite{black2024pi_0, intelligence2025pi_}, we employ Paligemma-3B~\cite{beyer2024paligemma} as the VLM backbone, Paligemma-300M as the action head.
We implement our motion head as a light-weight Diffusion Transformer (DiT)~\cite{peebles2023scalable} with 400M parameters.
We use 8 NVIDIA H200 GPUs for our experiments.
For pretraining, we use a batch size of 128, set the warm-up phase for 40k steps, and the joint training phase for another 100k steps.
For joint learning, we set $k$ to match the control frequency of each dataset, aligning the temporal window of both action and motion signals for the next second.
We use a batch size of 32 for finetuning on simulation benchmarks, with 30k steps on the LIBERO benchmark and 60k steps on the RoboTwin benchmark, with a learning rate of $5\mathrm{e}{-5}$ and a cosine annealing schedule.

\input{tabs/robotwin}
\subsection{LIBERO Benchmark}
\noindent \textbf{Setup.} We evaluate our method on the LIBERO benchmark~\cite{liu2023libero}, a large-scale simulation suite for evaluating generalist robot manipulation policies under diverse conditions. LIBERO consists of four suites: 1) Spatial: varying scene layouts and object positions, testing spatial generalization; 2) Object: different object instances and categories, testing object-level transfer; 3) Goal: varying goal configurations, evaluating semantic goal understanding and grounding, and 4) Long: long-horizon, multi-stage manipulation tasks, assessing temporal reasoning and compositional planning.
Each suite contains 10 manipulation tasks, and each task provides 50 expert demonstrations in simulation.
Each policy is tested for 50 rollouts per task, and we report success rates for each suite.

\noindent \textbf{Results.}
Tab.~\ref{tab:libero} shows that our approach further enhances the $\pi$-series VLAs, achieving an average success rate of $94.7\%$ with joint-learned $\pi_0$ and $97.5\%$ with joint-learned $\pi_{0.5}$, outperforming other baselines.
These results demonstrate that incorporating motion image diffusion provides complementary supervision that strengthens motion reasoning capabilities in large-scale VLA models.
Particularly, on the most challenging LIBERO-Long suite, which requires long-horizon planning and motion reasoning, we yield a performance gain by 4.0\% with $\pi_{0.5}$.
In contrast, models without explicit motion reasoning tend to produce abrupt or inconsistent action sequences when facing complex multi-stage tasks.
Compared to FlowVLA, which introduces flow images via visual Chain-of-Thought (CoT) reasoning~\cite{wei2022chain} only for pretraining, our joint training approach integrates motion learning directly into the policy optimization loop.
This end-to-end coupling between action and motion enables the model to align visuomotor representations with underlying motion reasoning, rather than relying solely on visual priors.
Furthermore, unlike WorldVLA~\cite{cen2025worldvla}, which combines future image prediction with action generation, our framework focuses on learning compact motion dynamics through diffusion in latent space, resulting in better efficiency and stronger generalization.
Overall, our results highlight that joint learning with motion image diffusion plays a critical role in enhancing large-scale VLA models to perform more stably and effectively across varying environments and manipulation complexities.

\subsection{RoboTwin Benchmark}
\noindent \textbf{Setup}. 
To evaluate the effectiveness of our method in bimanual manipulation, we conduct experiments on the RoboTwin 2.0 benchmark~\cite{chen2025robotwin}.
We follow the official setup using the dual Aloha-AgileX robot arm as our robot platform.
We select seven representative tasks that span diverse manipulation skills: \emph{Beat Block}, \emph{Click Alarmclock}, \emph{Move Can Pot}, \emph{Move Playingcard}, \emph{Place Shoe}, \emph{Scan Object}, \emph{Stacks Blocks Two}.
Each task contains 50 expert demonstrations used for training.
For evaluation, we adopt two settings in the benchmark: an easy mode with in-domain clean layouts, and a hard mode with domain randomization, including variations in textures, lighting, and background.
Each policy is evaluated over 100 trials per mode.
We compare our approach against representative imitation-based baselines, including Diffusion Policy (DP), RDT, and ACT, following the reported results from the official RoboTwin leaderboard.
DP, RDT, and ACT are trained per-task, following the official setting, whereas we train a single multi-task model across all seven tasks for $\pi_0$ and our method, highlighting their generalist learning capability.

\begin{figure}[t]
    \centering
    \includegraphics[width=1.0\linewidth]{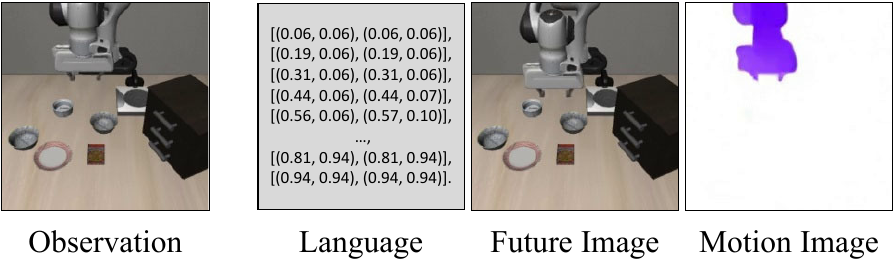}
    \caption{\textbf{Overview of different motion representations.} We compare three motion representations for joint learning, where the motion image is most effective.}
    \label{fig:representations}
\end{figure}
\input{tabs/representations}
\noindent \textbf{Results.} 
As shown in Tab.~\ref{tab:robotwin}, our method based on $\pi_0$ achieves the highest overall performance, with an average success rate of $58.0\%$.
Our method shows a substantial improvement over $\pi_{0}$ by $13.1\%$, and consistently improves success rates across both easy and hard modes, demonstrating stronger robustness to domain randomization and lighting variations.
In particular, tasks that involve long-horizon temporal coordination, such as \emph{Stack Blocks Two} and \emph{Move Playingcard}, show the most significant performance gains, suggesting that motion reasoning contributes to better temporal consistency and control stability.
Moreover, even though the motion head is not used during inference, the unified learning provides effective dense supervision that regularizes the VLM backbone, enabling the policy to learn richer temporal dependencies from limited demonstrations.
This improvement highlights the ability of our method to enhance multi-task bimanual robotic manipulation without modifying the original VLA architecture and inference pipeline.

\subsection{In-Depth Ablation Analysis}
\label{sec:ablation}
\begin{figure*}[t]
    \centering
    \includegraphics[width=1.0\linewidth]{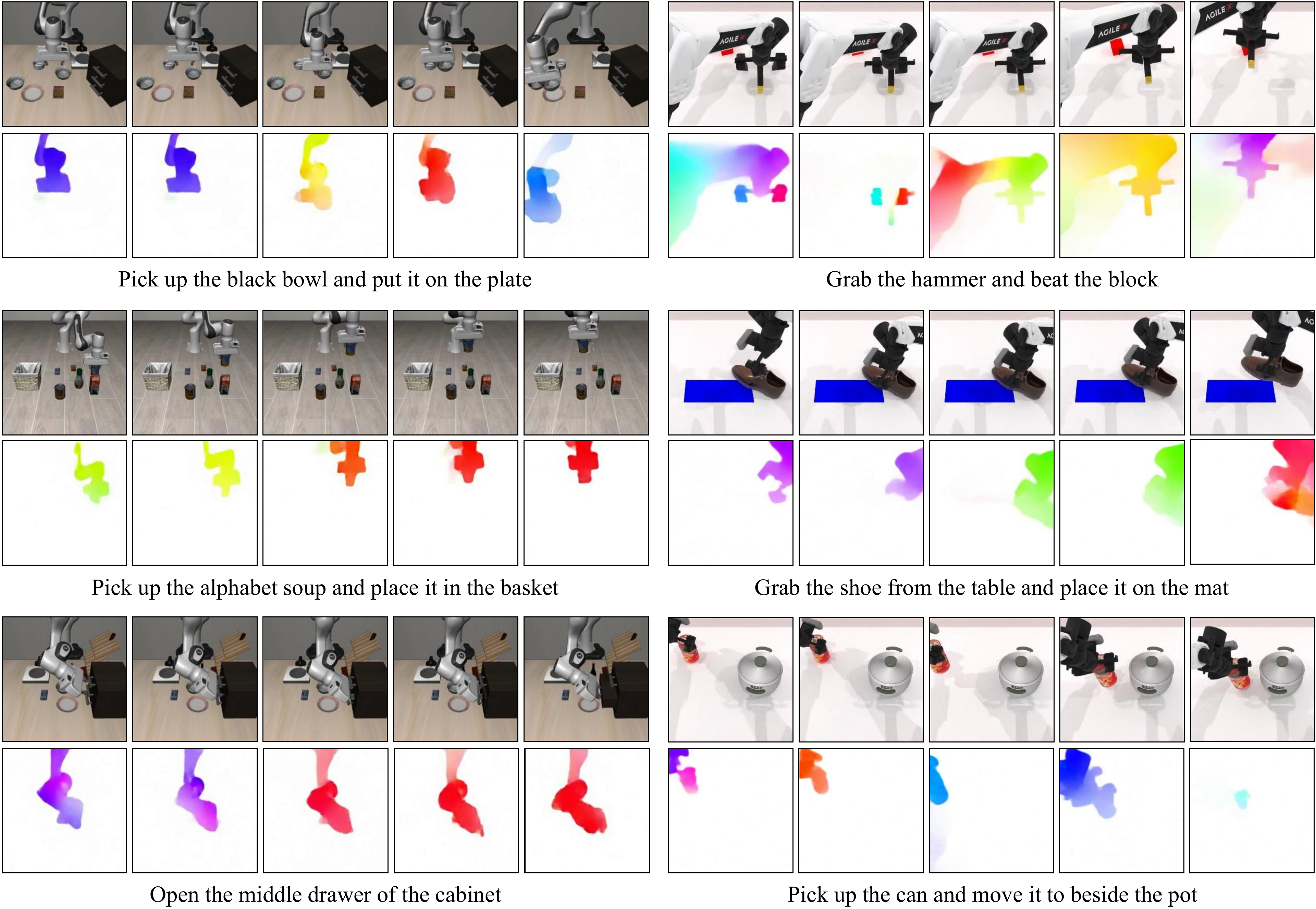}
    \caption{\textbf{Qualitative visualization of predicted action and motion during rollout.} We show example rollouts from both LIBERO and RoboTwin. For each example, the first row shows the observed frames with the robot actions predicted by the action head, and the bottom row visualizes the flow images predicted by the motion head. Please see supplementary materials for examples of long-horizon tasks.}
    \label{fig:exp_rollout}
\end{figure*}

\noindent \textbf{Investigation of Motion Representations.} 
We comprehensively investigate various motion representations to study their effect on joint learning in large-scale VLA models.
Specifically, as shown in Fig.~\ref{fig:representations}, we experiment with three popular types of motion representation: language, future image, and motion image. 
1) Language: Inspired by motion reasoning traces in~\cite{li2024llara,li2025hamster,yang2025magma}, we describe optical flow in natural language and learn it in an LLM style.
Concretely, we uniformly sample $8 \times 8$ pixels from the observation and extract their optical flow vectors.
Each pixel location is normalized to $[0, 1]$, and each flow is expressed as a location pair describing pixel displacement.
We modify $\pi$-series VLAs to predict these motion descriptions in parallel with action prediction, following an LLM-style modeling objective.
2) Future image: We represent motion implicitly by future images that describe the desired future appearance of the scene, inspired by recent approaches based on world models~\cite{cen2025worldvla}.
3) Motion image (Ours): We represent motion explicitly using optical-flow-based motion images that describe pixel-wise movement between frames.
For future image and motion image predictions, we employ the same dual-head architecture as described in Sec.~\ref{sec:method}, where the motion head learns in parallel with the action head.

In Tab.~\ref{tab:exp_representations}, we report the results of enhancing $\pi$-series VLAs with joint learning on the LIBERO benchmark.
Among all the representations, joint learning with motion images achieves the highest average success rate across all suites.
In contrast, language-based motion substantially underperforms other representations. 
Because language provides only discrete and low-frequency supervision, which is inherently inefficient and sparsely supervised, it lacks the spatial continuity to align textual motion descriptions with underlying pixel-space dynamics.
Although future image shows decent performance on short-horizon suites, it degrades performance on long-horizon tasks, as it emphasizes global visual appearance rather than explicit modeling of local motion dynamics.
This limitation becomes more pronounced in multi-stage scenarios, where reasoning about fine-grained motion is critical.
Meanwhile, motion image provides a dense and physically grounded representation that provides direct correspondence between observed motion and robot control, allowing the shared VLM backbone to learn temporally coherent visuomotor embeddings.
This ultimately improves the overall visuomotor reasoning ability of large-scale VLA models.

\begin{figure}[t]
    \centering
    \includegraphics[width=0.9\linewidth]{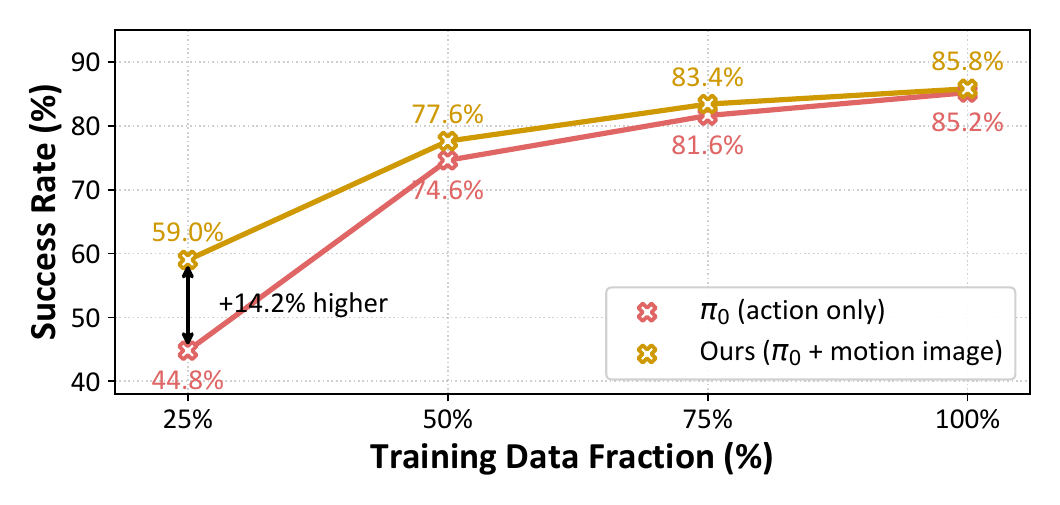}
    \caption{\textbf{Data efficiency on LIBERO-10.} We demonstrate that joint learning with motion image diffusion improves data efficiency over action-only learning.}
    \label{fig:exp_efficiency}
\end{figure}

\begin{figure*}[t]
  \centering
   \includegraphics[width=1.0\linewidth]{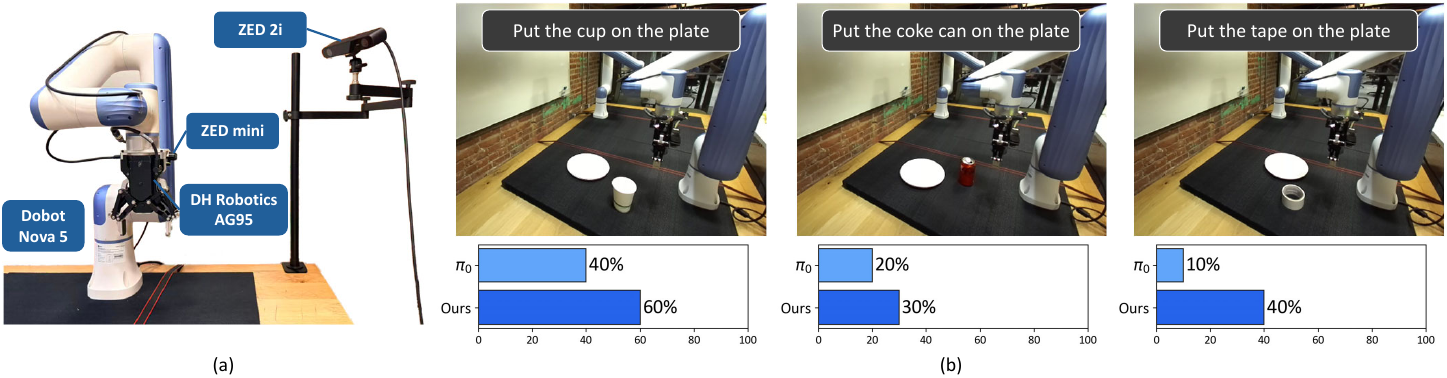}
   \caption{\textbf{Real-world experiments.} (a) Our setup. (b) Evaluation results across tabletop tasks under office scenarios.}
   \label{fig:exp_realworld}
\end{figure*}

\noindent \textbf{Relationship between Predicted Action and Motion.}
Fig.~\ref{fig:exp_rollout} presents qualitative visualizations of the predicted actions and motion images generated by our model during rollouts on both the LIBERO and RoboTwin benchmarks.
Each example shows an execution trajectory, where the first row shows the observed frames with the robot actions predicted by the action head, and the bottom row visualizes the corresponding optical-flow-based motion images predicted by the motion head.
Across various manipulation tasks, e.g., grasping, placing, and tool use, our model accurately predicts the direction and magnitude of robot arm and object movements, producing temporally smooth and spatially coherent flow fields.
Notably, the predicted motions align well with the underlying physical dynamics of each task.
For example, the predicted flows show consistent forward–backward movements during “Grab the hammer and beat the block”, and smooth translational motion during “Open the middle drawer of the cabinet”.
These results show that our motion image diffusion captures fine-grained motion cues at the pixel level, as additional dense supervision for VLA learning.
The tight correspondence between motion and control suggests that the model learns to reason about physical dynamics rather than mimicking expert trajectories, validating the benefit of our joint learning strategy.

\noindent \textbf{Data Efficiency.}
To evaluate the data efficiency of our method, we conduct experiments with $\pi_0$ on the LIBERO-Long benchmark, as shown in Tab.~\ref{fig:exp_efficiency}.
We subsample 25\%, 50\%, and 75\% of the full LIBERO dataset to assess performance under limited data regimes.
Overall, our approach consistently outperforms the baseline that only includes action learning, demonstrating stronger data efficiency.
Specifically, despite the challenge of this long-horizon suite, our joint-learned $\pi_0$ reaches a success rate of 59.0\% using only 25\% data, which is a notable improvement of 14.2\% over the $\pi_0$ baseline trained with the same data amount.
These results indicate that joint learning enhances the action learning ability of VLA models by incorporating motion understanding, thereby improving success rates from only a small amount of robotic data.

\subsection{Real-world Experiments}
\noindent \textbf{Setup.}
To further evaluate the effectiveness of our approach in real-world robotic manipulation, we conduct experiments using a Dobot Nova 5 6-DOF robot arm and a DH Robotics AG95 gripper. 
Our setup includes a two-camera configuration: a ZED 2i stereo camera providing a third-person view of the workspace, and a ZED Mini camera mounted on the wrist, offering an egocentric view.
We design three tabletop manipulation tasks representative of common office scenarios, and collect a small training dataset with 30 teleoperated demonstrations per task.
We conduct our experiments under a limited-data regime to simulate realistic low-data conditions.
We apply LoRA fine-tuning~\cite{hu2022lora} with a batch size of 32 on top of our joint-learned $\pi_0$.
We report the success rates and compare our results against the $\pi_0$ baseline.

\noindent \textbf{Results.}
As shown in Fig.~\ref{fig:exp_realworld}, our method consistently outperforms the $\pi_0$ baseline across all the tasks.
This aligns with the performance observed in the simulation benchmarks.
Specifically, our method achieves an average success rate of 43\%, improving $\pi_0$ by 23\%.
This demonstrates that incorporating motion image diffusion effectively enhances motion reasoning and physical consistency in robotic control, allowing the policy to better generalize from limited demonstrations.
This further highlights the effectiveness and robustness of our joint learning strategy under real-world conditions.

\section{Limitations}
Despite incorporating multi-view inputs, our motion head relies primarily on third-person observations, which makes it sensitive to occlusions and viewpoint biases in the third-person view.
Extending our method to predict and fuse the motion from multiple camera views would be an interesting direction for future work.
On the other hand, scaling action-free motion data with our strategy and exploring further improvements in VLA models remains a promising avenue for additional research.

\section{Conclusion}
In this work, we present joint learning with motion image diffusion, a novel strategy that augments Vision-Language-Action (VLA) models with explicit motion reasoning capabilities.
While existing VLAs directly map multimodal observations and instructions to actions, they often just imitate expert trajectories.
Our method bridges this gap through joint learning with motion image diffusion: a dual-head architecture that the action head predicts action chunks as in standard VLAs, a motion head implemented as a Diffusion Transformer (DiT) predicts optical-flow-based motion images to capture future dynamics.
Through joint optimization, the shared VLM backbone learns representations that couple control with motion evolution.
This design enhances motion-aware understanding without altering the inference pathway, thus preserving real-time performance.
Extensive experiments in both simulation and real-world environments show our success rates reaching 97.5\% on LIBERO and 58.0\% on RoboTwin, validating the effectiveness of motion image diffusion for advancing large-scale VLA models.

{
    \small
    \bibliographystyle{ieeenat_fullname}
    \bibliography{main}
}

\input{sec/X_suppl}

\end{document}

%% file: tabs/libero.tex
\begin{table*}[t]
  \centering
  \small
  \setlength{\tabcolsep}{17.8pt}
  \begin{tabular}{>{\centering\arraybackslash}p{4cm} >{\centering\arraybackslash}p{1.2cm} >{\centering\arraybackslash}p{1.2cm} >{\centering\arraybackslash}p{1.2cm} >{\centering\arraybackslash}p{1.2cm} @{\hspace{1cm}}>{\centering\arraybackslash}p{1.2cm}}
    \toprule
    \textbf{Model} & \textbf{Spatial} & \textbf{Object} & \textbf{Goal} & \textbf{Long} & \textbf{Average} \\
    \midrule
    DP~\cite{chi2025diffusion} & 78.3\% & 92.5\% & 68.3\% & 50.5\% & 72.4\% \\
    Octo~\cite{team2024octo} & 78.9\% & 85.7\% & 84.6\% & 51.1\% & 75.1\% \\
    OpenVLA~\cite{kim2024openvla} & 84.7\% & 88.4\% & 79.2\% & 53.7\% & 76.5\% \\
    SpatialVLA~\cite{qu2025spatialvla} & 88.2\% & 89.9\% & 78.6\% & 55.5\% & 78.1\% \\
    WorldVLA~\cite{cen2025worldvla} & 87.6\% & 96.2\% & 83.4\% & 60.0\% & 81.8\% \\
    FlowVLA~\cite{zhong2025flowvla} & 93.2\% & 95.0\% & 91.6\% & 72.6\% & 88.1\% \\
    $\pi_{0}\text{-FAST}$~\cite{pertsch2025fast} & 96.4\% & 96.8\% & 88.6\% & 60.2\% & 85.5\% \\
    $\pi_{0}$~\cite{black2024pi_0} & 96.8\% & \cellcolor{tab2}98.8\% & 95.8\% & 85.2\% & 94.2\% \\
    $\pi_{0.5}$~\cite{intelligence2025pi_} & \cellcolor{tab1}98.8\% & \cellcolor{tab3}98.2\% & \cellcolor{tab1}98.0\% & \cellcolor{tab2}92.4\% & \cellcolor{tab2}96.9\% \\
    \midrule
    Ours ($\pi_{0}$) & \cellcolor{tab2}98.6\% & 97.8\% & \cellcolor{tab2}96.4\% & \cellcolor{tab3}85.8\% & \cellcolor{tab3}94.7\% \\
    Ours ($\pi_{0.5}$) &  \cellcolor{tab3}98.4\% & \cellcolor{tab1}99.2\% & \cellcolor{tab3}96.0\% & \cellcolor{tab1}96.2\% & \cellcolor{tab1}97.5\%\\
    \bottomrule
  \end{tabular}
  \caption{\textbf{Success rates on the LIBERO benchmark across four suites.} We rank the best performances with \sethlcolor{tab1}\hl{red} (1st), \sethlcolor{tab2}\hl{orange} (2nd), and \sethlcolor{tab3}\hl{yellow} (3rd) highlights. Our joint learning strategy further improves $\pi_0$ and $\pi_{0.5}$ to achieve the top performance over other methods.}
  \label{tab:libero}
\end{table*}

%% file: tabs/robotwin.tex
\begin{table*}[t]
  \centering
  \small
  \setlength{\tabcolsep}{7.5pt}
  \begin{tabular}{
    >{\centering\arraybackslash}p{2.4cm}||
    >{\centering\arraybackslash}p{0.8cm}
    >{\centering\arraybackslash}p{0.8cm}
    >{\centering\arraybackslash}p{0.8cm}
    >{\centering\arraybackslash}p{0.8cm}
    >{\centering\arraybackslash}p{1.35cm}|
    >{\centering\arraybackslash}p{0.8cm}
    >{\centering\arraybackslash}p{0.8cm}
    >{\centering\arraybackslash}p{0.8cm}
    >{\centering\arraybackslash}p{0.8cm}
    >{\centering\arraybackslash}p{1.35cm}
  }
    \toprule
    \multirow{2}{*}{\textbf{Task}} &
    \multicolumn{5}{c|}{\textbf{Easy}} &
    \multicolumn{5}{c}{\textbf{Hard}} \\
    \cmidrule(lr){2-11}
    & DP~\cite{chi2025diffusion} & RDT~\cite{liu2024rdt} & ACT~\cite{zhao2023learning} & $\pi_0$~\cite{black2024pi_0} & \textbf{Ours ($\pi_0$)} 
    & DP~\cite{chi2025diffusion} & RDT~\cite{liu2024rdt} & ACT~\cite{zhao2023learning} & $\pi_0$~\cite{black2024pi_0} & \textbf{Ours ($\pi_0$)} \\
    \midrule
    Beat Block & 42\% & \textbf{77\%} & 56\% & 51\% & 67\% & 0\% & \textbf{37\%} & 3\% & 6\% & 16\% \\
    Click Alarmclock & 61\% & 61\% & 32\% & \textbf{79\%} & 65\% & 5\% & 12\% & 4\% & 18\% & \textbf{25\%} \\
    Move Can Pot & 39\% & 25\% & 22\% & 43\% & \textbf{81\%} & 0\% & \textbf{12\%} & 4\% & 3\% & 2\% \\
    Move Playingcard & 47\% & 43\% & 36\% & 42\% & \textbf{57\%} & 0\% & 11\% & 0\% & \textbf{16\%} & 3\% \\
    Place Shoe & 23\% & 35\% & 5\% & 31\% & \textbf{43\%} & 0\% & 7\% & 0\% & 11\% & \textbf{13\%} \\
    Scan Object & 9\% & 4\% & 2\% & 19\% & \textbf{27\%} & 0\% & 1\% & 0\% & 3\% & \textbf{6\%} \\
    Stack Blocks Two & 7\% & 21\% & 25\% & 51\% & \textbf{66\%} & 0\% & 2\% & 0\% & \textbf{5\%} & 0\% \\
    \midrule
    \textbf{Average} & 32.6\% & 38.0\% & 25.4\% & 45.1\% & \textbf{58.0\%} & 0.7\% & \textbf{11.7\%} & 1.6\% & 8.9\% & 9.3\% \\
    \bottomrule
  \end{tabular}
  \caption{\textbf{Results on the RoboTwin benchmark.} Comparison of success rates (\%) across seven tasks under \textbf{Easy} (in-domain) and \textbf{Hard} (domain-randomized) settings. Our joint learning strategy achieves the highest overall performance, improving $\pi_0$ by +12.9\% on Easy and +0.4\% on Hard settings on average.}
  \label{tab:robotwin}
\end{table*}

%% file: tabs/representations.tex
\begin{table}[t]
  \centering
  \small
  \begin{tabular}{>{\centering\arraybackslash}p{1cm}>{\centering\arraybackslash}p{3cm}>{\centering\arraybackslash}p{2cm}}
    \toprule
    \textbf{Model} & \textbf{Supervision} & \textbf{Average} \\
    \midrule
    \multirow{4}{*}{$\pi_0$}
      & action only   & 94.2\% \\
      & action + language      & 86.1\% \\
      & action + future image  & 93.6\% \\
      & action + motion image  & \textbf{94.7\%} \\
    \midrule
    \multirow{4}{*}{$\pi_{0.5}$}
      & action only   & 96.9\% \\
      & action + language      & 95.1\% \\
      & action + future image  & 95.7\% \\
      & action + motion image  & \textbf{97.5\%} \\
    \bottomrule
  \end{tabular}
  \caption{\textbf{Investigation of joint learning with different motion representations on the LIBERO benchmark.} In Sec.~\ref{sec:ablation}, we compare the effectiveness of representing motion as language descriptions, future images, and motion images for joint learning. Motion image yields the best overall performance.}
  \label{tab:exp_representations}
\end{table}

%% file: sec/X_suppl.tex
\clearpage
\setcounter{page}{1}

\appendix

\section{Additional Implementation Details}
\subsection{Flow Image for Motion Supervision}
As described in Sec.~\ref {sec:training}, we use image-based optical flow (termed as flow image) for our motion supervision.
Here, we provide additional details on how flow images are constructed for training.

Given an image observation $o_t \in \mathbb{R}^{3\times H \times W}$ from the training dataset, we compute the optical flow between $o_t$ and $o_{t+k}$ using RAFT~\cite{teed2020raft}.
RAFT outputs a Cartesian optical flow field $F=(f_x, f_y) \in \mathbb{R}^{2 \times H \times W}$, where $f_x$ and $f_y$ denote horizontal and vertical flow components, respectively.
To obtain a visually interpretable flow image, we first convert optical flow into a polar representation.
We compute the flow magnitude with $\sqrt{f_x^2 + f_y^2}$ and normalize it by a factor of 64, which is the dataset-wide maximum magnitude to ensure consistent scaling across samples.
We map the flow angle $\alpha=\operatorname{atan2}(f_y, f_x)$ to the hue channel of an HSV representation, and map the magnitude to the saturation channel.
The value channel is set to a constant maximum intensity of $1$ to ensure high-contrast visualization.
Eventually, we convert the HSV tensor into an RGB image via color space transformation.
This produces a flow image that encodes motion direction and magnitude in a smooth and meaningful color space.
Representing optical flow in RGB format also allows motion supervision to align naturally with the visual processing pipeline of the VLM backbone, improving generalizability and compatibility with diffusion-based generation.

\begin{figure*}[t!]
    \centering
    \includegraphics[width=1.0\linewidth]{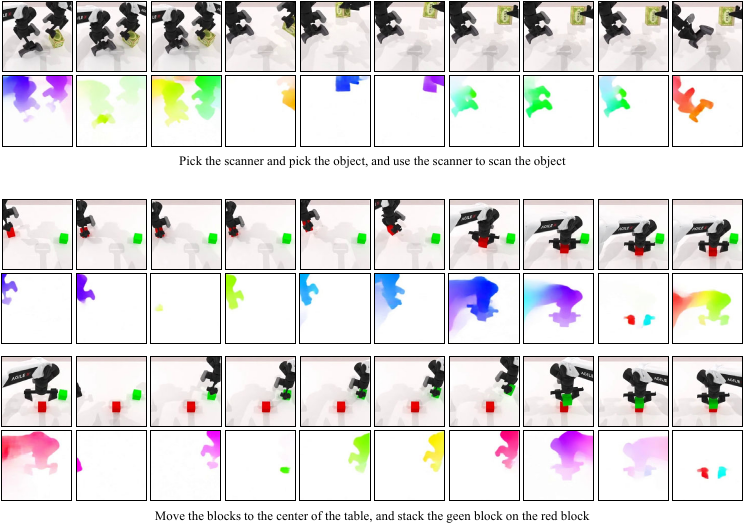}
    \caption{\textbf{Qualitative results of long-horizon tasks in the RoboTwin benchmark.} Our method maintains consistent performance in bimanual long-horizon tasks. The predicted flow images remain stable and exhibit clear motion patterns that align with task progress.}
    \label{fig:sup_robotwin}
\end{figure*}

\begin{figure*}[t!]
    \centering
    \includegraphics[width=0.92\linewidth]{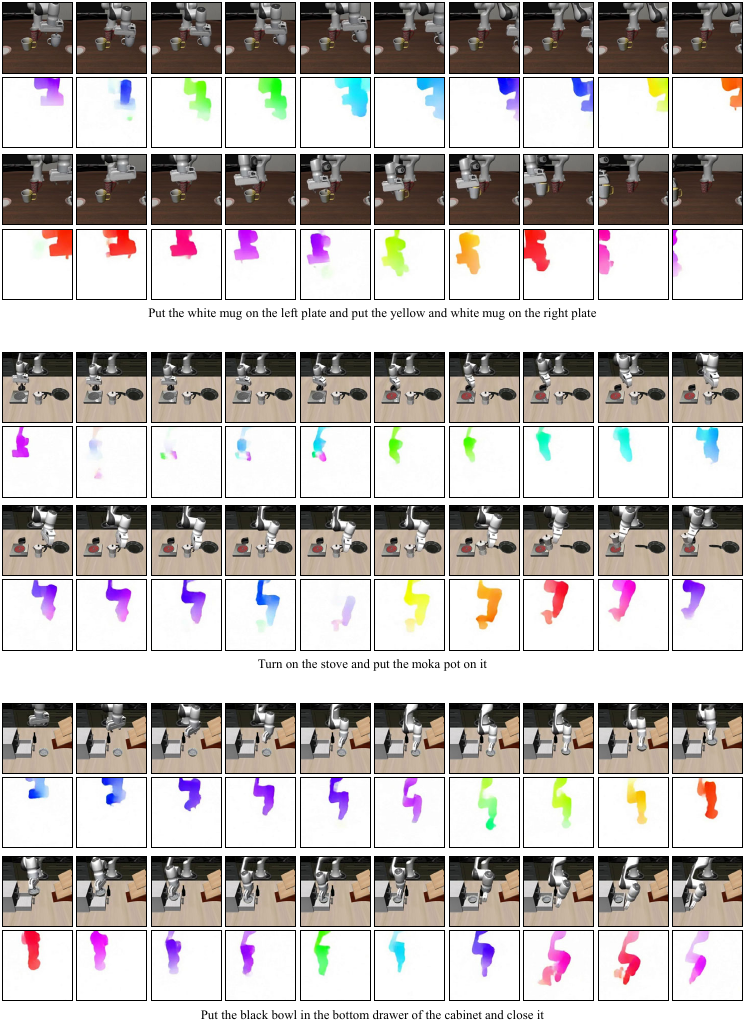}
    \caption{\textbf{Qualitative results of long-horizon tasks in the LIBERO benchmark.} Our method maintains consistent performance in long-horizon tasks. The motion head accurately captures fine-grained dynamics, e.g., the subtle rotational motion to turn the stove knob.}
    \label{fig:sup_libero}
\end{figure*}

\begin{figure*}[t!]
    \centering
    \includegraphics[width=0.9\textwidth]{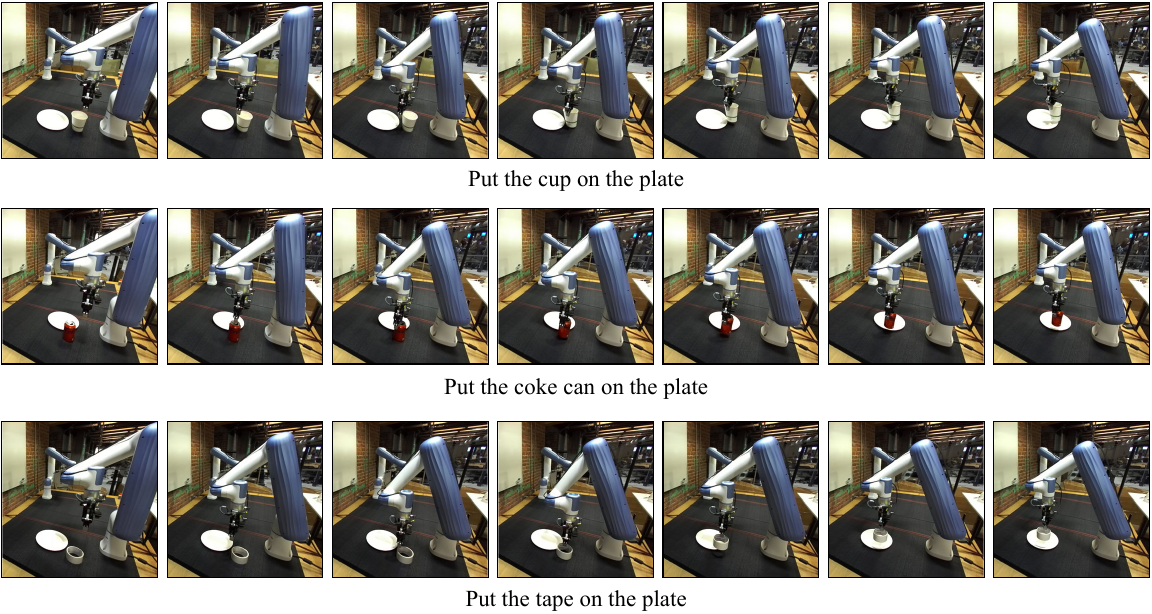}
    \captionof{figure}{\textbf{Qualitative results of real-world experiments.} Our policy produces smooth and goal-directed motions across the tasks, demonstrating reliable visuomotor control in real-world settings.}
    \label{fig:sup_real}
\end{figure*}

\begin{figure*}[b!]
    \centering
    \includegraphics[width=0.9\textwidth]{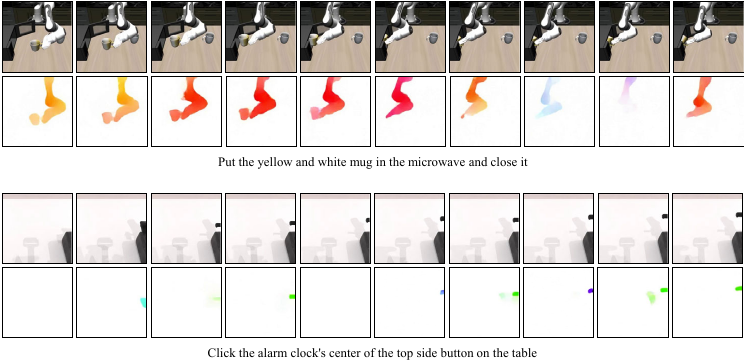}
    \captionof{figure}{\textbf{Examples of our failure cases.} These examples indicate that failures often come from third-person view occlusions. Incorporating multi-view motion prediction across cameras is a promising direction for future work.}
    \label{fig:sup_failure}
\end{figure*}

\begin{figure*}[t]
    \centering
    \includegraphics[width=0.95\linewidth]{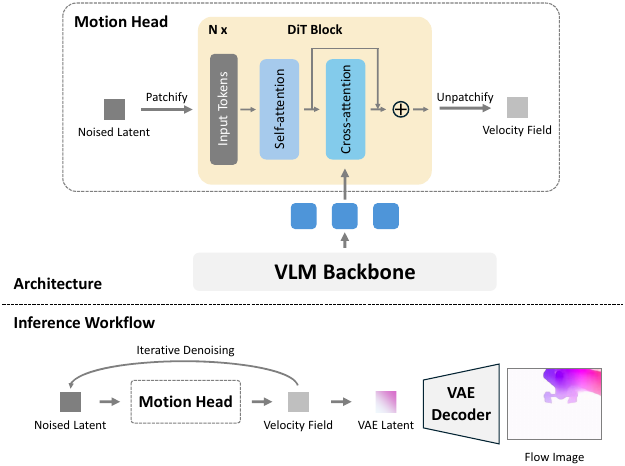}
    \caption{
    \textbf{Detailed architecture and inference workflow of the motion head.} 
    \textbf{(a) Architecture}: We implement the motion head as a Diffusion Transformer (DiT)~\cite{peebles2023scalable} that operates on VAE latents of flow images. Each DiT block applies self-attention over latent tokens and cross-attention conditioned on the VLM prefix. 
    \textbf{(b) Inference Workflow}: At inference time, the motion head iteratively denoises a noised latent into a clean VAE latent, which is then decoded into a flow image via the VAE decoder without any optimization.
    }
    \label{fig:sup_motionhead}
\end{figure*}

\subsection{Architecture}
Our architecture consists of a shared VLM backbone and two lightweight heads: an action head and a motion head.
We use the same VLM backbone (PaliGemma-3B) and the action head (PaliGemma-300M) as in $\pi_0$~\cite{black2024pi_0} and $\pi_{0.5}$\cite{intelligence2025pi_}, and leverage their pretrained checkpoints for our development.
These pretrained components remain unchanged during the integration of the motion head.
The VLM backbone encodes the observation $o_t$ and instruction $l$ into a multimodal prefix (denoted as $z_t$ in the paper).
The action head generates an action chunk $A_t$ conditioned on this prefix.

As shown in Fig.~\ref{fig:sup_motionhead}, we implement the motion head as a Diffusion Transformer (DiT)~\cite{peebles2023scalable}, which operates on VAE latents of flow images.
During training, each ground-truth flow image of size $3 \times 224 \times 224$ is encoded into a latent tensor $m_t$ of shape $4 \times 28 \times 28$ using the frozen SDXL-VAE encoder.
We then perform diffusion on these latent tokens.
The core of DiT is a stack of $N=6$ standard Transformer blocks.
At each diffusion timestep, a noisy latent $m_t^\tau$ is processed by the blocks.
Each block applies self-attention over latent tokens and cross-attention conditioned on the VLM prefix $z_t$, where latents act as queries and $z_t$ as keys and values.
This design allows the motion head to inject multimodal context into the denoising process while keeping the module lightweight.
The DiT predicts a velocity field $v(m_t^\tau, o_t)$ required by the flow-matching objective.
At inference time, we obtain a clean latent from Gaussian noise, which is then decoded using the frozen VAE decoder to produce a flow image.

\section{Additional Qualitative Results}
\subsection{Long-Horizon Simulation Evaluation}
We provide additional qualitative results of our method on long-horizon manipulation tasks in the simulation benchmarks.
Fig.~\ref{fig:sup_robotwin} and Fig.~\ref{fig:sup_libero} present rollouts from the RoboTwin and LIBERO-Long benchmarks, further demonstrating that our method maintains consistent performance in long-horizon settings.
Across both benchmarks, the predicted flow images remain stable and exhibit clear motion patterns that align with task progress, similar to the results observed in short-horizon suites.
Notably, the motion head accurately captures fine-grained dynamics, e.g., the subtle rotational motion to turn the stove knob in “Turn on the stove and put the moka pot on it” in Fig.~\ref{fig:sup_libero}. 
This validates that motion image diffusion provides reliable pixel-level motion cues that support long-range visuomotor reasoning.

\subsection{Real-world Evaluation}
For real-world experiments, we record observations from two camera viewpoints (third-person ZED 2i and wrist-mounted ZED Mini), along with absolute joint angles at a rate of 15Hz.
We finetune and deploy a single model across all three tabletop tasks (“Put the cup on the plate”, “Put the coke can on the plate”, and “Put the tape on the plate”).
Fig.~\ref{fig:sup_real} presents the rollouts from our real-world experiments.
The policy produces smooth and goal-directed motions that successfully place objects onto the target plate across different initial configurations.

\subsection{Failure Cases}
We present representative failure cases in Fig.~\ref{fig:sup_failure}.
Although our VLM backbone inherits the multi-view observation capability of $\pi$-series VLAs~\cite{black2024pi_0, intelligence2025pi_}, the motion head primarily relies on the third-person observation.
As a result, this leads to failures when critical scene information is occluded from the third-person camera.
In the first example (``Put the yellow and white mug in the microwave and close it''), the cup becomes stuck inside the microwave -- an event not visible from the third-person viewpoint, preventing the model from correct motion predictions.
In the second example (``Click the alarm clock's center of the top side button on the table''), the alarm clock is only partially visible, leading to incomplete motion cues.
These cases indicate that incorporating multi-view motion prediction and fusing motion cues across multiple cameras is a promising direction for future work.